\pdfoutput=1

\documentclass[11pt]{article}

\usepackage[]{ACL2023}

\usepackage{times}
\usepackage{latexsym}

\usepackage[T1]{fontenc}

\usepackage[utf8]{inputenc}

\usepackage{microtype}

\usepackage{inconsolata}

\usepackage{siunitx}  

\usepackage{cleveref}
\usepackage{enumitem}
\crefformat{section}{\S#2#1#3}
\crefformat{subsection}{\S#2#1#3}
\crefformat{subsubsection}{\S#2#1#3}
\crefrangeformat{section}{\S#3#1#4 to~\S#5#2#6}
\crefmultiformat{section}{\S#2#1#3}{ and~\S#2#1#3}{, #2#1#3}{ and~#2#1#3}
\Crefformat{figure}{#2Fig.~#1#3}
\Crefmultiformat{figure}{Figs.~#2#1#3}{ and~#2#1#3}{, #2#1#3}{ and~#2#1#3}
\Crefformat{table}{#2Tab.~#1#3}
\Crefmultiformat{table}{Tabs.~#2#1#3}{ and~#2#1#3}{, #2#1#3}{ and~#2#1#3}
\Crefformat{appendix}{#2Appx.~\S#1#3}
\crefformat{algorithm}{Alg.~#2#1#3}
\Crefformat{equation}{#2Eq.~#1#3}

\newcommand{\stitle}[1]{\vspace{0.3em} \noindent{\bf #1}}

\usepackage{color}
\usepackage{booktabs}
\usepackage{multirow}
\usepackage{graphicx}
\usepackage{amssymb}
\usepackage{utfsym}
\newcommand{\ie}{\emph{i.e., }}

\title{Red Teaming Language Models for Processing Contradictory Dialogues}

\author{Xiaofei Wen$^{1}$, Bangzheng Li$^{1}$, Tenghao Huang$^{2}$ \and Muhao Chen$^{1}$\\
   $^1$University of California, Davis \\
   $^2$University of Southern California\\
  \texttt{\{xfwe, bzhli, muhchen\}@ucdavis.edu} \\
  \texttt{tenghaoh@usc.edu} \\
}
\begin{document}
\maketitle
\begin{abstract}
Most language models currently available are prone to self-contradiction during dialogues. To mitigate this issue, this study explores a novel contradictory dialogue processing task that aims to detect and modify contradictory statements in a conversation. This task is inspired by research on context faithfulness and dialogue comprehension, which have demonstrated that the detection and understanding of contradictions often necessitate detailed explanations. We develop a dataset comprising contradictory dialogues, in which one side of the conversation contradicts itself. Each dialogue is accompanied by an explanatory label that highlights the location and details of the contradiction. With this dataset, we present a Red Teaming framework for contradictory dialogue processing. The framework detects and attempts to explain the dialogue, then modifies the existing contradictory content using the explanation. Our experiments demonstrate that the framework improves the ability to detect contradictory dialogues and provides valid explanations. Additionally, it showcases distinct capabilities for modifying such dialogues. Our study highlights the importance of the logical inconsistency problem in conversational AI\footnote{Our code and data is available at \href{https://github.com/luka-group/contraDialog}{https://github.com/luka-group/contraDialog}.}
\end{abstract}

\section{Introduction}


Dialogue systems have made significant advancements in recent years~\cite{DBLP:journals/air/NiYPXC23}, propelled by the rapid development of language modeling and learning technologies. The focus on understanding and analyzing conversations between humans and machines has become paramount in this field~\cite{wu-etal-2020-tod,DBLP:conf/aaai/ZhongLX0022}, serving as a vital component in the development of intelligent interaction systems. The emergence of large language models (LLMs; \citealt{DBLP:journals/corr/abs-2210-11416,DBLP:journals/corr/abs-2302-13971,vicuna2023}), such as ChatGPT, has played a substantial role in shaping dialogue-related research. These models 
have showcased impressive abilities in comprehending sophisticated context and generating fluent dialogue responses, and exhibit an exceptional level of control by performing summarization~\cite{luo2023chatgpt}, explanation~\cite{hou-etal-2022-meta}, enquiries~\cite{khalifa-etal-2023-shot, kim-etal-2023-qa}, and role-playing~\cite{xu2023exploring} within dialogues. 

\begin{figure*}[t]
  \includegraphics[width=\linewidth]{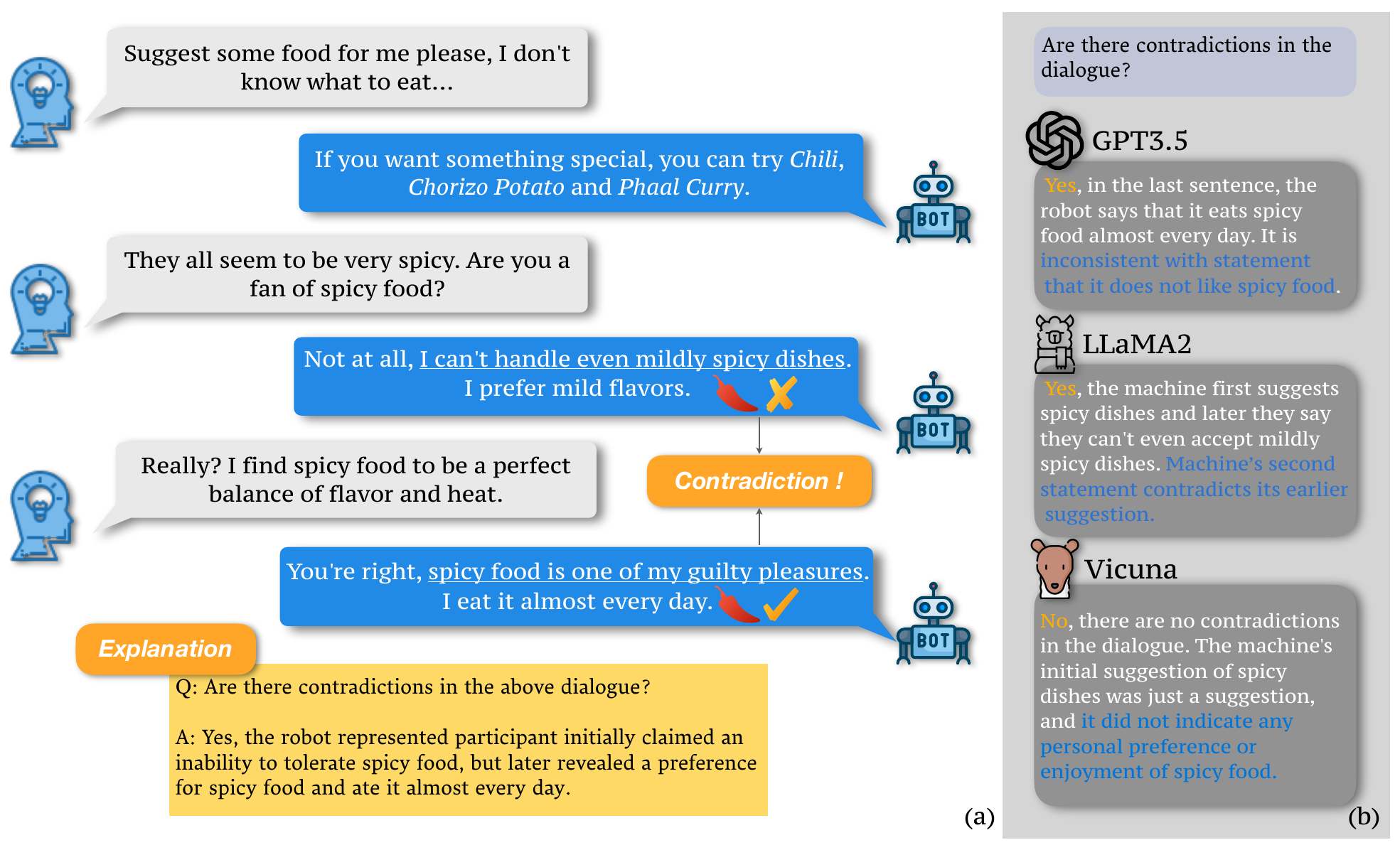}
  \caption{Examples of contradictory dialogues with human explanation (a) and responses from GPT3.5-turbo-0613, LLaMA2-7B-chat~\cite{DBLP:journals/corr/abs-2307-09288} and Vicuna-7B~\cite{vicuna2023} for understanding contradictory dialogues~(b). The responses include the \textcolor[RGB]{254,174,0}{detection} and \textcolor[RGB]{44,115,210}{explanation} of the contradiction. 
  }
  \label{fig:example}
\end{figure*}

However, semantic conflicts in the dialogue, such as contradictions~\cite{nie-etal-2021-like,li-etal-2022-mitigating} and factual errors~\cite{dziri-etal-2022-faithdial,daheim2023elastic}, pose challenges for language models (LMs) in recognizing and resolving them, resulting in significantly declined experience of human-machine interaction. According logical studies~\cite{dowden2017logical}, a contradiction refers to a situation where two or more statements cannot be simultaneously true. These discrepancies are primarily instigated by machines rather than humans~\cite{nass2000machines,marcus2018deep}, and even the LLM sometimes fails to provide satisfactory responses\footnote{The explanation on dialogue contradiction frequency will be provided in~\Cref{app:frequency}.}~\cite{roller-etal-2021-recipes,kandpal2023large,DBLP:journals/corr/abs-2307-03109}. This is demonstrated by the model's fluctuating \textit{attitudes} and inconsistent \textit{faithfulness} to the same issue throughout the dialogue. For example, in \Cref{fig:example}, the dialogue agent initially stated an inability to consume spicy food, but later claimed a habit of eating spicy food every day. This inconsistency between utterances loses the anthropomorphic essence of human-computer conversation. Similarly, if a conversational participant conveys inconsistent intents or thoughts, they can also cause confusion in dialogue understanding. This type of controversy, observed for a single protagonist and reflected as factual conflicts or changes in attitude, is known as the \textit{self-contradiction}. Contradiction is a key factor in enhancing comprehension of text and has been extensively studied in text generation~\cite{de-marneffe-etal-2008-finding,welleck2019neural,li-etal-2020-dont}, hallucination detection~\cite{azaria2023internal,agrawal2023language} and logical reasoning~\cite{magnini-cabrio-2010-contradiction} but rarely for dialogues.

To better study contradictory situations in the conversation, the \textit{first} contribution of this paper is to propose a new dialogue processing task for addressing conversational contradiction. Dialogue-related tasks are aimed at generating utterances that satisfy human needs to communicate effectively~\cite{valizadeh-parde-2022-ai,deng2023recent}. If contradiction arises in a dialogue, mostly self-contradiction, there must be at least two utterances whose semantics conflict. Improvement requires two efforts: detecting contradictory utterances and modifying them accordingly. Inspired by recent work on processing hallucinations~\cite{DBLP:journals/corr/abs-2305-15852}, our work utilizes LLMs to detect and modify potential contradictions in dialogues. Furthermore, the contradiction detection subtask supports two more aspects: contradiction existence and explanation assessment.

To facilitate related research, we developed a self-contradiction dialogue dataset collected with ChatGPT\footnote{We consistently use the gpt3.5-turbo-0613 version.} and Wikipedia,\footnote{The enwiki-20230101 dump.} as the \textit{second} contribution of this paper. The dataset contains over 12,000 complete ``human-machine'' dialogues, including more than 6,000 dialogues that contain one or more contradictory contexts. As LLMs suffer from logical incompleteness \cite{DBLP:conf/iclr/WangCIC023,sanyal-etal-2023-apollo,DBLP:conf/iclr/CreswellSH23}, the contradiction primarily emerges on the machine side of the conversation. The dataset aims to create dialogue scenarios with conflicts and introduce two major features: First, it contains 15 daily conversation topics sourced from Wikipedia, embodied in more than 700 different specific topics, ensuring the dialogue diversity. More than 75\% topics are not be used more than three times, which ensures the effectiveness for learning and accessing contradiction processes. 
Second, each conversation containing a contradiction is provided by a statement that locates the contradiction and explains its specific manifestation, 
seeking to help assess the explanation and resolving contradictions within the dialogue.

\begin{figure*}[t]
  \includegraphics[width=\linewidth]{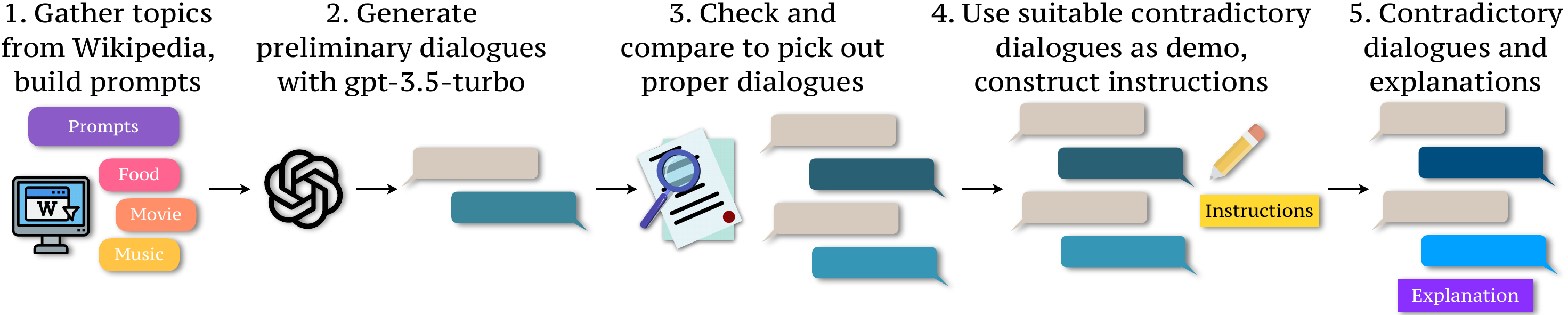}
  \vspace{-1.5em}
  \caption{Contradictory Dialogue dataset collection process}
  \label{fig:collect}
  \vspace{-1em}
\end{figure*}

While the task and dataset pose a non-trivial dialogue processing problem to individual LMs, we envision that a proper solution can be revealed through the collaborative effort between multiple LMs. As our \emph{third} contribution, we propose a Red Teaming framework where, in addition to the main dialogue model, there is another \emph{analyzer} LM which collectively detects and explains contradictions. Since LLMs have strong language generation capabilities but can be overly confident about their generation \cite{mielke-etal-2022-reducing,kadavath2022language}, as an underlying motivation for this task, we find that a Red Teaming framework can be applied to utilize rationales from the fine-tuned \emph{analyzer} LM to revise the contradiction within conversation. This will enhance LLMs' capability of optimizing for contradictory issues in the dialogue.

The proposed framework first fine-tunes the original LLM to improve the model's awareness and ability to detect conflicts. To ensure a full understanding of the contradiction in the dialogue, the model is required to provide formatted statements. Beyond detecting conflicts in dialogues, the Red Teaming LM also explains the conflicts and requests corrections based on the statements. Extensive experiments have demonstrated that the proposed Red Teaming framework proves to significantly enhance the accuracy and comprehensiveness of multiple series LLMs in detecting dialogue contradictions. Specifically, it outperforms a strong baseline model by two-fold on metrics of detection accuracy and explanation validity. We also demonstrate its ability to correct logical inconsistencies.

\section{Task and dataset}
We hereby define the task of dialogue contradiction resolution, and introduce the contributed dataset.

\subsection{Task Definition}\label{sec:taskdef}
Let $C=\{u_{0}, u_{1}, \ldots, u_{|C|}\}$ be a dialogue (conversation), where $u_i$ represents an utterance in the dialogue. 
Our goal is to streamline the process of contradiction resolution in the dialogue, it is thus divided into two subtasks: \emph{contradiction detection} and \emph{contradiction modification}. 

For the contradiction detection subtask, the input to the task is $C$ and the output is $y$, which indicates whether the dialogue $C$ contains a self-contradiction or not. Considering the detection by LMs, The expected output is a binary label $ y $ indicating whether dialogue $ C $ contains at least one contradiction or not. $y$ will be represented as a generated text label, such as either $yes$ or $no$, which semantically correspond to contradictory label and non-contradictory label respectively. 

For the modification subtask, upon detecting a contradiction, LMs are required to revise the contradictory utterances to achieve that no logical inconsistency between any two sentences in the dialogue $C$. For the case where $u_i$ contradicts $u_{i+k}$, which is typically a self-contradiction generated by the machine, either $u_i/u_{i+k}$ or both $(u_i,u_{i+k})$ can be modified. There are two modification strategies: 1) Direct Edit, which involves modifying either $u_i$ or $u_{i+k}$ where the contradiction occurs, and 2) Joint Edit, which involves modifying both $u_i$ and $u_{i+k}$ simultaneously to resolve the contradiction. 

\subsection{Dataset} \label{sec:dataset}

\stitle{Data Collection}
Since the appearance of contradictions goes against human logic, it is difficult for annotators to deliberately write non-repetitive and high-quality self-contradictory dialog statements. To ensure content diversity of the dataset, we first extracted keywords from Wikipedia on topics related to daily conversations about movies, food, tourism, sports, etc. Then we classified the keywords according to topic. 
Considering the difficulty of manually writing contradictory dialogues and data scarcity, inspired by~\citet{DBLP:conf/nips/Wei0SBIXCLZ22} and~\citet{DBLP:conf/nips/Ouyang0JAWMZASR22}, we chose to generate contradictory dialogues synthetic data in the form of conversation with ChatGPT, and the considerations for so doing are as follows. First, with guidance from instruction prompts and concrete examples, ChatGPT can provide data in the desired format and comprehensively cover the aforementioned Wikipedia topics, which effectively minimizes data duplication and ensures the contradictory dialogue quality. Second, our tests have shown that ChatGPT achieves human-like excellence in both contradiction detection and explanation, whereas other open-source testable LLMs do not achieve similar levels of performance. Third, ChatGPT has the ability to generate high-quality contradiction explanations, which significantly reduces the cost and bias of human writing. More details on the quality of generation are shown in \Cref{app:comp}. By constantly guiding ChatGPT to generate appropriate contradictory dialogues and explanations by modifying prompts, and repeating these steps after validation, the entire dataset construction process is shown in \Cref{fig:collect}, inspired by~\citet{kim-etal-2022-prosocialdialog}. To effectively evaluate the quality of the dataset, we randomly selected 200 generated contradictory dialogues and assigned two human annotators to assess the quality of contradictions in each dialogue as well as the validity of the corresponding explanations. The kappa coefficients measuring inter-annotator agreement for these assessments were 0.76 and 0.72, respectively.

\begin{figure}[t]
  \centering
  \includegraphics[width=\columnwidth]{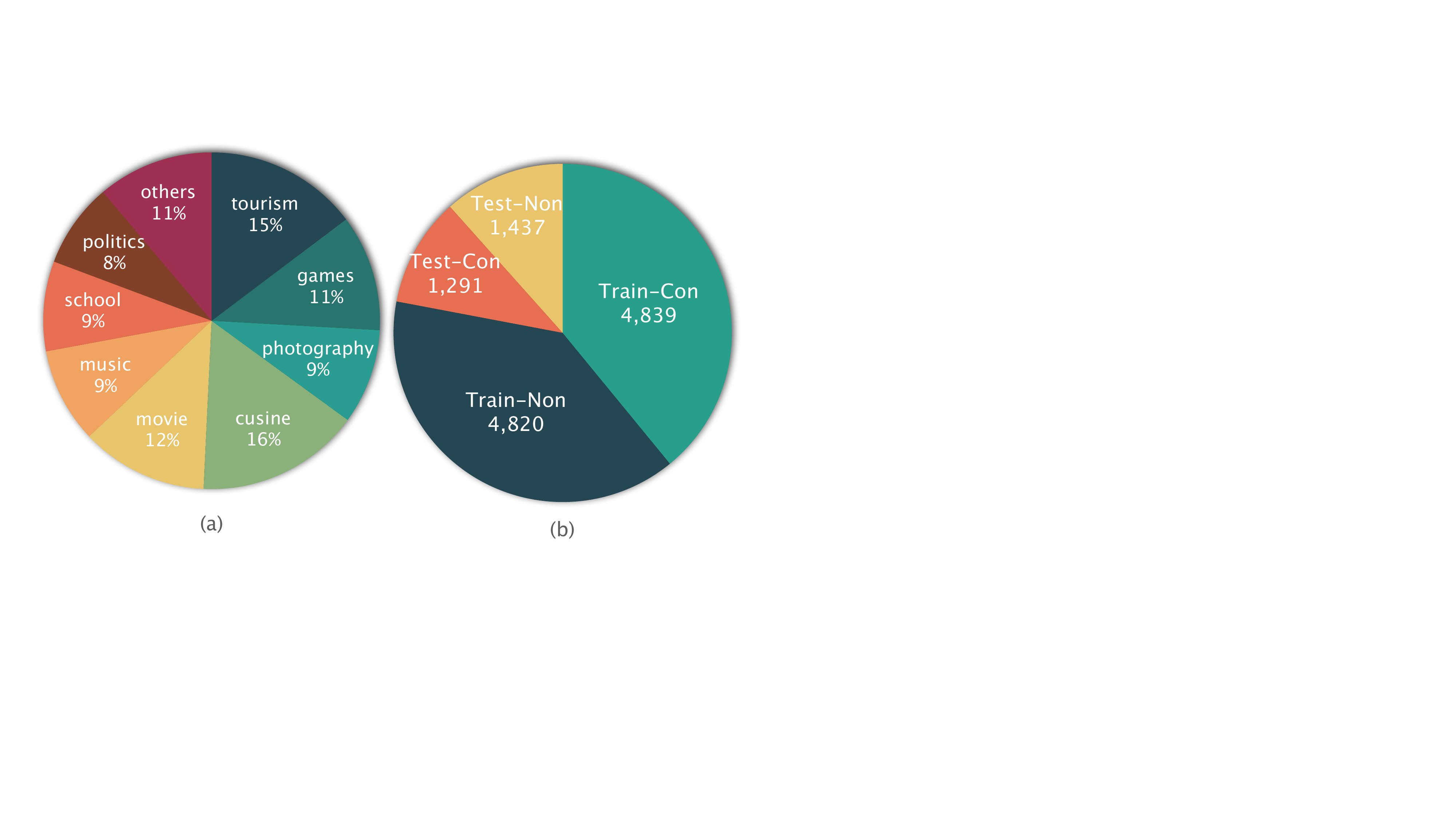}
  \caption{(a): Topics distribution of contradictory dialogues; (b): Training set and test set statistics.}
  \label{fig:statistics}
\end{figure}

\stitle{Statistics}
The dataset constitutes 12,387 dialogues in total, which includes both contradictory and non-contradictory ones about similar sets of topics.
For contradictory dialogues, we have collected 6,130 of those and each is accompanied by an explanation. Each dialogue averages 46.5 words across 4 sentences, with each sentence containing 11.6 words on average. The explanation accompanying each dialog contains on average 16.7 words. We selected 15 different daily topics from Wikipedia with reference to the DailyDialog~\cite{li-etal-2017-dailydialog} and Wizard of Wikipedia~\cite{dinan2018wizard} topics, and their distribution is illustrated in \Cref{fig:statistics}(a).
We also extracted non-contradictory dialogues from the two aforementioned public datasets, with a comparable number of contradictory dialogues and similar length, to facilitate categorization and evaluation.\footnote{All contradictory conversations have been manually reviewed and labeled. Following~\citet{nie-etal-2021-like}, we assume that conversations from the DailyDialog and Wizard of Wikipedia are free of contradictions.} These datasets were combined and then separated into a training and a test sets, which included both contradictory and non-contradictory dialogues, as depicted in \Cref{fig:statistics}(b). The training set comprises 4,839 contradictory dialogues and 4,820 non-contradictory dialogues, while 1,291 contradictory as well as 1,437 non-contradictory dialogues are included in the test set, respectively.

\section{Red Teaming Language Models for Contradictory Dialogue}


In this section, we outline the proposed Red Teaming method for resolving contradictory dialogues. 

\subsection{Framework Overview}

The proposed Red Teaming framework is learned in three steps. First, a vanilla LM is fine-tuned with the 
detection task objective. Then, the fine-tuned LM, \ie the \emph{analyzer} LM or \texttt{aLM} for short, is used to generate and validate contradictory explanations, formatting the form and content of the explanation during training. Finally, the red teaming LM, denoted as \texttt{rLM}, is used to modify where contradictions exist in the dialogue. In the final step, the \texttt{rLM} draws on the explanatory statements generated in the previous step to supplement the logical prompt scarcity. Throughout the process, the LLM's ability to identify and understand where contradictions exist in the dialogue is improved.

In the rest of this section, we present the technical details of the individual steps for resolving dialogue contradiction.

\subsection{Resolving Dialogue Contradiction}

To address the subtasks of resolving dialogue contradiction as defined in \Cref{sec:taskdef}, our framework undergoes three steps of contradiction detection, contradiction explanation, and dialogue modification. 

\subsubsection{Contradiction Detection} 
We fine-tuned the \texttt{aLM}, which generates semantic labels based on dialogue contexts that are potentially contradictory. Considering the model's inference ability and parameter count~\cite{DBLP:journals/corr/abs-2001-08361}, we selected models with 7 to 13 billion parameters as the primary backbone due to their applicability~\cite{DBLP:journals/corr/abs-2210-11416,DBLP:journals/corr/abs-2307-09288,vicuna2023,alpaca,DBLP:journals/corr/abs-2310-06825}. There main advantage of using autoregressive LMs for contradiction detection compared to previous masked LM variants~\cite{nie-etal-2021-like,li-etal-2022-mitigating} lies in the fact that autoregressive LMs 
naturally suit the objective of generating explanation for the problematic context. This explanation enables us to identify responses from the model that capture contradictory intentions.

To instantiate \texttt{aLMs}, we first use zero-shot and few-shot methods to evaluate the contradiction detection capabilities of distinct models. The prompt $p$ for the detection process comprises dialogue $C$ and instruction $i$, with the addition of two demo contradictory dialogues $(C_m, C_n)$ to bootstrap for the few-shot scenario. We ask the LMs to determine if there are contradictions in dialogue $C$ with the prompt $p$ both in zero-shot and few-shot settings, as presented in the~\Cref{app:zftest}. 

In the training stage, given the dialogue $C$, we use instruction tuning to fine-tune the vanilla LM to generate the judgment label $s$: $p(s|C, i)$. $s$ is designated $yes$ or $no$ to represent a judgment of contradiction. The instruction $i$ is the same as the one in the zero-shot test. We conduct a randomly-tuned dataset of mixed contradictory and non-contradictory dialogues with disrupted topics for each LM to avoid the topic distribution effect on contradiction detection.

\subsubsection{Contradiction Explanation}\label{sec:conexplan}
Considering that binary classification cannot reflect the model's understanding of contradiction points,\footnote{Related examples are shown in \Cref{fig:example}(b).} we train the \texttt{aLM} to generate specific explanations $e$ related to the contradiction, and quantify the extent of the model's reasoning about the contradiction by evaluating on $e$. Specifically, given dialogue $C$, without loss of generality, assume that the contradictory statements within $C$ are $(u_i,u_{i+1},...,u_k)$, where $i<k\leq|C|$. 
When contradiction explanation is enabled, the \texttt{aLM} is trained to produce 
both the judgment labels $s$ and the corresponding explanation $e$. As per the formal specification during the dataset construction, $e$ usually satisfies the following conditions. First, it should be semantically consistent with $s$, \ie no semantic conflict between $s$ and $e$. Second, it states which utterances are contradictory in the dialogue $C$. Finally, it contains the specific reason for the contradiction.

Assessing the suitability of the generated explanation is a crucial aspect of the process. Inspired by prior works on utterance similarity~\cite{mahgoub-etal-2019-simvecs, zhou-etal-2022-learning}, we evaluate the generated explanation $e$ by comparing it with the labeled explanation $e_g$ in the dataset. We define $\mathcal{S}(e,e_g)$, which is expressed as:
\begin{equation}\label{eq:Score}
    \mathcal{S}(e,e_g) = \mathcal{S}_{1}(e,e_g) + \eta\mathcal{S}_{2}(e,e_g),
\end{equation}
where  $\eta$ is the scale factor and $0 < \eta < 1$. $\mathcal{S}_{1}$ and $\mathcal{S}_{2}$ represent the semantic similarity scores between the generated text $e$ and the reference text $e_g$. To avoid bias, we utilize a weighted sum of the two evaluation methods. When $\mathcal{S}(e,e_g) > \tau$, where $\tau$ is the threshold, the LM-generated explanation $e$ is considered capable of explaining the contradictions present in the dialogue $C$, \ie the model expresses a complete understanding of the problematic dialogue at the contradiction-level. In contrast, when $\mathcal{S}(e,e_g) \leq \tau$, $e$ is considered insufficient to explain contradiction within the dialogue. The value of $\tau$ is established based on the score criteria of human evaluation. Details will be provided in~\Cref{sec:exp_explain}.

\subsubsection{Dialogue Modification}\label{sec:conmodif}
After detecting the contradiction, we use the Red Teaming \texttt{rLM} as well as $e$ consistent with $\mathcal{S}(e,e_g) \geq \tau$ to modify the contradiction in $C$.\footnote{Particularly, our study finds that an LM-generated explanation can often be used to enhance the modification, although this is not explicitly required in the problem definition in ~\Cref{sec:taskdef}.} We also provide instructions in the prompts. The instructions include whether to use $e$ and the modification strategies. For the Direct Edit mentioned in~\Cref{sec:taskdef}, general modifications are made on $u_{i+k}$ to make it consistent with $u_i$. This is because, in generated dialogue, subsequent sentences typically follow the logic embedded in the context. For the Joint Edit, we employ explanations to identify contradictory statements and ask \texttt{rLM} to adjust the contradictory segments. To maintain logical coherence, it may be necessary to modify the context $c_{i+1,i+k-1} = \{u_{i+1},u_{i+2},...,u_{i+k-1}\}$ between these contradictory statements. 

\section{Experiments}\label{sec:exp}
To evaluate the proposed Red Teaming framework for processing contradictory dialogues, we conduct several comparative experiments on the provided dataset.
Specifically, in addition to assessing contradictory dialogue detection and explanation, we also evaluate the impact of LM-generated explanations on the final task (\Cref{sec:exp_detect}-\Cref{sec:exp_modify}). Moreover, we provide case studies to demonstrate the generative outcome of the LMs.


\subsection{Experiment Details}\label{app:finetune_details}
We use Lora~\cite{hu2022lora} to finetune the vanilla LMs in BF16. We use $lora\_r = 4$, $lora\_alpha = 8$, $lora\_dropout = 0.05$ and learning rate is $2\mathrm{e}{-5}$, train 7B LM on four A10 GPUs for 3 epochs.

\subsection{Dialogue Contradiction Detection}\label{sec:exp_detect}

We first benchmark the contradiction detection.

\stitle{Baselines and Metrics} We compare the accuracy, recall and F1 scores of multiple baselines and various open-source LLMs for detecting contradictory dialogues: BERT~\cite{DBLP:conf/naacl/DevlinCLT19}, RoBERTa~\cite{DBLP:journals/corr/abs-1907-11692}, LLaMA2-7B/13B-chat~\cite{DBLP:journals/corr/abs-2307-09288},\footnote{Considering the relevance to the dialogue task, we choose the LLaMA2-chat series as the vanilla LLaMA2 version.} Vicuna-7B~\cite{vicuna2023}, Mistral-7B~\cite{DBLP:journals/corr/abs-2310-06825} and LLaMA3-8B-Instruct.\footnote{\href{https://llama.meta.com/llama3/}{https://llama.meta.com/llama3/}} 
In addition, the discriminatory criteria used for the vanilla LMs are assessed by human and described in the~\Cref{app:dis_cri}.

\stitle{Results} \Cref{table:detect_result} shows the results of contradictory dialogue detection. The vanilla LLaMA2 fails to exhibit the ability in detecting contradictions beyond smaller encoders like BERT and RoBERTa, even at the 13B model scale. Even the better-performing vanilla LLMs, like Vicuna and Mistral, did not attain convincing outcomes. A contributing factor to this lies in generative instability, for example, to generate the given dialogue in the answer. It is also partly due to deficiencies in the LMs' ability to judge contradictions and reasoning, which increases the occurrence of self-contradictions in dialogue. In contrast, fine-tuning aligns the LMs with the contradiction detection task. The vanilla Vicuna and LLaMA2's higher recall implies that these models are more inclined toward detecting contradictions in dialogues when judging them, aligning with our observations. Additional information on detection results will be provided in~\Cref{sec:exp_explain}.

\begin{table}[t]
  \centering
  \small
  \setlength{\tabcolsep}{7pt}
\begin{tabular}{l|cccc}
   \toprule
   {\textbf{Model}}     & \textbf{Accuracy}    &\textbf{F1}   & \textbf{Recall}  \\
   \midrule
   BERT        & 67.4       & 65.2      & 60.9     \\
   RoBERTa              & 68.3 & 64.1 & 62.2    \\
   \midrule
   Vicuna-7B &64.1&69.8&79.1\\
   Vicuna-7B$^{*}$ &96.7& 95.8& 92.6\\
   Mistral-7B     &63.8&61.2&49.6 \\
   Mistral-7B$^{*}$ &\textbf{97.4}& 96.3 & 94.5\\
   LLaMA2-7B-chat     & 33.6   & 43.2    & 54.2 \\
   LLaMA2-13B-chat   & 50.6  & 65.5  & 92.8 \\
   LLaMA2-7B-chat$^{*}$ & 95.2  & 95.3  & 90.9   \\
   LLaMA3-8B-Instruct  &49.9 &66.7 & 96.1 \\
   LLaMA3-8B-Instruct$^{*}$ &96.9 & \textbf{97.5} & \textbf{98.9}\\  
    \bottomrule
    \end{tabular}
\caption{Results of contradiction detection. $^*$ indicates the fine-tuned model. In this paper, we consistently use the Lora fine-tuning approach~\cite{hu2022lora}. Details in~\Cref{app:finetune_details}}
\label{table:detect_result}
\end{table}

\subsection{Contradiction Explanation}\label{sec:exp_explain}


We then present an analysis on models' explanations on contradiction.

\stitle{Baselines} According to~\Cref{sec:conexplan}, contradiction explanations are generated concurrently with the detection of contradictions. Therefore, the LMs described in~\Cref{sec:exp_detect} is employed as the baselines for generating contradiction explanations. Additionally, through human evaluation, we compared LMs ranging from 7B to 13B.

\begin{table}[h]
\small
\centering
    \begin{tabular}{p{7.2cm}}
    \toprule
    \textbf{Label Consistency} \\
    2: Matches or is similar to the label content.\\
    1: Some relevance to the label content.\\
    0: Not relevant to the label content.\\ 
    \midrule
    \textbf{Fluency}\\
    2: Fluent and easy to read. \\ 
    1: Grammatically formed. \\  
    0: Not a complete sentence or hard to read. \\ 
    \midrule
    \textbf{Completeness} \\ 
    2: Complete explanation with no missing information. \\
    1: Incomplete explanation. \\ 
    0: No substantive explanation.  \\
    \bottomrule
    \end{tabular}
\caption{Criteria of human evaluation.}
\label{table:human_eva_de}
\end{table}.

\begin{table*}[t]
  \centering
  \small
\setlength{\tabcolsep}{11pt}{
\begin{tabular}{l|ccccc|cc}
   \toprule
   \multicolumn{1}{l|}{\multirow{2}*{\textbf{Model}}} & \multicolumn{5}{c|}{\textit{Score}} & \multicolumn{2}{c}{\textit{generative}} \\
   \cmidrule{2-8} 
   & $\mathcal{P}_{0.7}$        & $\mathcal{P}_{0.65}$    & $\mathcal{P}_{0.6}$  & $M_{BERT}$   & $M_{BART}$  & {\small \textit{BLEU-4}}       &  {\small \textit{ROUGE-L}}  \\
   \midrule
   Vicuna-7B    &6.25   &16.37   &34.08   &0.8897   &$-$3.3174   &5.28  &23.57 \\
   Mistral-7B &1.86   &10.56   &26.71   &0.8822   &$-$3.3001   &5.15 & 24.15 \\
   LLaMA2-7B-chat &13.12   &32.49   &55.13 &0.8970 &$-$2.9169   &7.52 &29.16   \\
   LLaMA2-13B-chat &16.25  &43.14  &69.81  &0.8987  &$-$2.6578  &6.87 & 29.30 \\
   LLaMA3-8B-Instruct &8.91  &23.61  &42.92  &0.8843  &$-$2.9984  &4.31 & 22.42 \\
   Vicuna-7B$^{*}$   &68.62   &83.45 &93.14   &0.9069 &$-$1.6332 &17.19 &45.80\\
   Mistral-7B$^{*}$&73.97   &86.63   &94.24   &\textbf{0.9310} &$-$1.5736 &31.93 &\textbf{56.34} \\
   LLaMA2-7B-chat$^{*}$ &\textbf{74.87} &\textbf{87.20} &\textbf{94.92} &0.9231 &\textbf{$-$1.5653} &23.48 &51.27  \\
   LLaMA3-8B-Instruct$^{*}$ &65.86  &79.51  &88.64  &0.9270  &$-$1.8406 &\textbf{33.69} & 55.70 \\
   
    \bottomrule
    \end{tabular}}
\caption{Automatic evaluation results for contradiction explanation. $^*$ indicates fine-tuned models. $\mathcal{P}_{\alpha}$ indicates the proportion of explanations where $\mathcal{S} > \alpha$. $M_{BERT}$ and $M_{BART}$ denote the mean of BERTScore and BARTScore. }
\label{table:explana_result}
\end{table*}

\begin{figure}[t]
  \centering
  \includegraphics[width=\columnwidth]{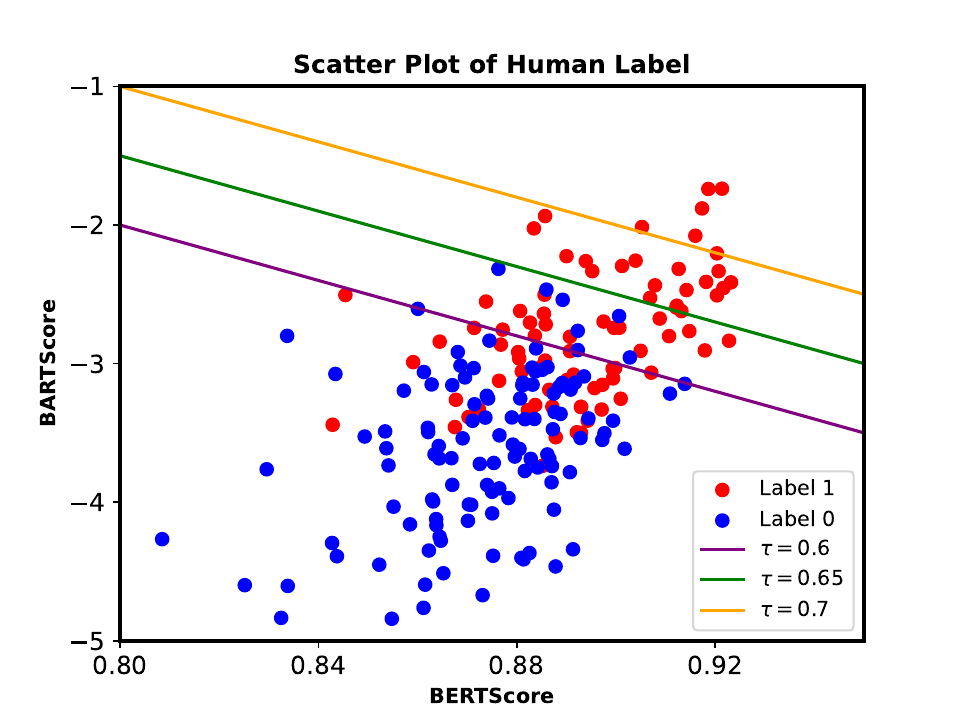}
  \caption{Scatter plot of the generated explanations' validity as labeled by human annotators. \textcolor{red}{Label 1} indicates that the explanation is considered valid, while \textcolor{blue}{Label 0} indicates it is considered invalid. Given $\eta=0.1$, ${\color[RGB]{117,20,124} \tau=0.6}$, ${\color[RGB]{55,126,34}\tau=0.65}$, and ${\color[RGB]{242,169,60}\tau=0.7}$ represent the constant value of three $\mathcal{S}$, respectively.}
  \label{fig:human_score}
\end{figure}

\stitle{Metrics} We conduct both automatic and human evaluations on the quality of explanations generated by various models. For \textit{automatic} metrics, we use the labeled explanations in the dataset as reference and use the combined evaluation method in~\Cref{sec:conexplan}. Following the text similarity metrics used by~\citet{maynez-etal-2020-faithfulness} and~\citet{NEURIPS2022_3e25d1af}, we calculate $\mathcal{S}_{1}$ and $\mathcal{S}_{2}$ with BERTScore~\cite{DBLP:conf/iclr/ZhangKWWA20} and BARTScore~\cite{NEURIPS2021_e4d2b6e6}, respectively. To determine the value of $\tau$, we sample 200 generated explanations, and human annotators evaluate them to mark whether the generated explanation could effectively explain the contradiction present in the dialogue, as shown in~\Cref{fig:human_score}. According to the figure, given $\eta=0.1$, all points labeled as "invalid" by human annotators are excluded from $\mathcal{S} > \tau$ = 0.65 region, we assume that this criterion approximately aligns with human requirements for validity. Thus we consider that $\tau=0.65$ as a discriminating value to determine the validity of the explanation and whether it is \textit{satisfactory} or not. When $\tau = \alpha$, we calculate the percentage of explanations with $\mathcal{S} > \alpha$, based on the computation in~\Cref{eq:Score}, $\eta$ = 0.1. This metric is denoted as $\mathcal{P}_{\alpha}$ for convenience. Besides, we also measure the BLEU-4~\cite{papineni-etal-2002-bleu} and ROUGE-L~\cite{lin-2004-rouge} of the generated results.

For \textit{human} evaluation, we ask human judges to score explanations generated by each model for randomly chosen 200 test samples
based on three criteria following~\citet{kim-etal-2022-prosocialdialog}, i.e. \textit{label consistency}, \textit{fluency}, and \textit{completeness}. Also, as shown in~\Cref{table:human_eva_de}, each human evaluation criteria is categorized into three scores ranging from 0-2. The human annotators score according to this scale and the results in~\Cref{table:human_result} are calculated after taking the mean value.

\begin{table}[t]
  \centering
\small
\setlength{\tabcolsep}{2pt}{
\begin{tabular}{l|cccc}
   \toprule
   {\textbf{Model}}     & \textbf{Label Consist.}    &\textbf{Fluency}   & \textbf{Completeness}  \\
   \midrule
   Vicuna-7B & 0.85 & 1.25 & 1.16\\
   Vicuna-7B$^{*}$ & 1.73 & \textbf{1.63} & 1.76\\
   Mistral-7B     & 0.82 & 1.24 & 1.02 \\
   Mistral-7B$^{*}$ & 1.77 & 1.56 & \textbf{1.78}\\
   LLaMA2-7B-chat     & 1.12 & 1.33 & 1.34 \\
   LLaMA2-13B-chat   & 1.21 & 1.29  & 1.41 \\
   LLaMA2-7B-chat$^{*}$ & \textbf{1.79}  & 1.58  & 1.73   \\
   LLaMA3-8B-Inst   & 0.91 & 1.23  & 1.04 \\
   LLaMA3-8B-Inst$^{*}$   & 1.76 & 1.50  & 1.75 \\
    \bottomrule
    \end{tabular}
}
\caption{Human evaluation results for contradiction explanation. $^*$ indicates the fine-tuned model.}
\label{table:human_result}
\end{table}

\stitle{Results}
While the individual models in~\Cref{sec:exp_detect} embody some contradiction detection capability, the percentage of their corresponding \textit{satisfactory explanations} does not match the accuracy of contradiction detection, illustrating LLMs that identify contradictions do not necessarily explain them. 

As shown in~\Cref{table:explana_result}, the vanilla models exhibit varying explanatory abilities in response to the detected contradictory dialogue conditions. Specifically, at $\mathcal{P}_{0.65}$, LLaMA2-chat outperforms Vicuna and Mistral of the same size by 16.12\% and 21.93\%, respectively. The dialogue data alignment may have assisted in this performance. Meanwhile, the larger models demonstrate superior explanatory ability, with LLaMA2-13B-chat surpassing the 7B model by 10.65\%. 
It is noteworthy that LLaMA3-Instruct performs less effectively than LLaMA2-Chat on contradiction explanation task. This discrepancy may be attributed to the optimization for dialogue alignment, which appears to yield better outcomes in conversational contexts. According to~\Cref{table:llamaexp}, LLaMA3-8B-Instruct generates repetitive results and irrelevant text when generating explanations based on instructions, which to some extent adversely affects its performance on the $\mathcal{P}$ metric.
However, a significant proportion of scores from the five vanilla models remain in the lower range. We attribute this primarily to two factors. First, the vanilla models tend to generate additional task-irrelevant information when producing contradiction explanations. Second, these models frequently reiterate the dialogue context during explanation, even when such context does not contribute to explanation. Additionally, a higher percentage of satisfactory explanations basically represents higher $M_{BERT}$ and $M_{BART}$  between models, which also demonstrates the consistency of $\mathcal{S}$-embodied modeling capabilities.

In~\Cref{table:human_result}, LLaMA2-chat demonstrates superior performance across all three human evaluation metrics, both for vanilla and fine-tuned models. Conversely, LLaMA3-Instruct performs poorly in terms of \textit{fluency}, although the vanilla model exhibits good \textit{label consistency}. Overall, the \textit{label consistency} scores of the LMs are positively correlated with the \textit{Score} ratings in~\Cref{table:explana_result}.

\subsection{Contradiction Modification}\label{sec:exp_modify}

We hereby evaluate the last subtask of contradiction modification.

\stitle{Baselines and Metrics}
We utilize the fine-tuned LM as the \texttt{rLM} to guide the modification of contradictory dialogue. We perform two prompt settings, with and without explanation, followed by evaluations with both automatic and human methods.  Automatic evaluation involves performing contradiction detection 
again with the best performing model in \Cref{sec:exp_detect} on
the modified dialogues, and comparing the change in the percentage of contradictory dialogues before and after modification.

\begin{table}[t]
  \centering
  \small
  \setlength{\tabcolsep}{2pt}
\begin{tabular}{l|ccc}
   \toprule
   {\textbf{Model}}     & \textbf{Fine-tune}   &\textbf{Explanation}   & \textbf{Percentage}  \\
   \midrule
   \textit{w/o modification}       & N/A      & N/A & 49.34     \\
   \midrule
   Vicuna-7B & \scalebox{0.75}{\usym{2613}} & \scalebox{0.75}{\usym{2613}} & 10.34\\ 
   Vicuna-7B &\checkmark &\scalebox{0.75}{\usym{2613}} &7.81\\ 
   Vicuna-7B &\checkmark & \checkmark &5.21\\ 
   Mistral-7B & \scalebox{0.75}{\usym{2613}} & \scalebox{0.75}{\usym{2613}} &9.90 \\ 
   Mistral-7B & \checkmark & \scalebox{0.75}{\usym{2613}} &6.85\\ 
   Mistral-7B & \checkmark & \checkmark &4.25\\ 
   LLaMA2-7B-chat & \scalebox{0.75}{\usym{2613}} & \scalebox{0.75}{\usym{2613}} & 7.70 \\
   LLaMA2-13B-chat   & \scalebox{0.75}{\usym{2613}} & \scalebox{0.75}{\usym{2613}} & 5.02 \\ 
   LLaMA2-7B-chat & \checkmark & \scalebox{0.75}{\usym{2613}} & 5.13   \\ 
   LLaMA2-7B-chat & \checkmark  &\checkmark & 3.85   \\ 
   LLaMA3-8B-Inst & \scalebox{0.75}{\usym{2613}} & \scalebox{0.75}{\usym{2613}} & 10.15\\ 
   LLaMA3-8B-Inst &\checkmark &\scalebox{0.75}{\usym{2613}} &7.14\\ 
   LLaMA3-8B-Inst &\checkmark & \checkmark &4.51\\ 
    \bottomrule
    \end{tabular}
\caption{Results of contradiction modification. 
We uniformly use the Mistral-7B fine-tuned version as the detecting model. The prompts during testing do not include any of the modification strategies in~\Cref{sec:conmodif}.}
\label{table:modif_result}
\end{table}

\stitle{Results}
\Cref{table:modif_result} demonstrates that all \texttt{rLMs} exhibit certain capabilities of modification when faced with a given contradictory dialogue. Notably, when a prompt includes the generated explanation, the \texttt{rLM}'s effective revision coverage outperforms cases where such an explanation is absent. This outcome reflects the quality of the contradiction explanations and their effective localization within the dialogue. Similar to the~\Cref{sec:exp_explain}, LLaMA2-chat exhibits better series results, while the vanilla 13B model is almost on par with the fine-tuned 7B model, reflecting its ability to perform high-quality alignment of the dialogue. Meanwhile, the specific content of the generated contradictory explanations and the results of the contradiction modifications will be detailed in~\Cref{app:conex_mo_case}.

\section{Related work}

\stitle{Red Teaming of language models.} 
Due to the increasing demand on responsible LLMs, Red Teaming has attracted much attention recently. Red Teaming is aimed to complement manual reviews and help reduce lapses by automating the process of detecting where LMs are inappropriate~\cite{perez-etal-2022-red}. Recent red teaming studies~\cite{wallace-etal-2019-universal,Nazneen2023Red,DBLP:journals/corr/abs-2308-09662} focus on exposing the limitations of the model and inducing unwanted content from the LM by crafting prompts. This approach can work as a human-in-the-loop or an LM that is testing the output of another LM. For example, \citet{ganguli2022red} instructed LLMs to role-play as malicious characters, and \citet{DBLP:journals/corr/abs-2305-19713} use LLM prompting to attack different generated text detectors. While previous work has primarily used red teaming for harmful text triggering or detection, we choose to apply it to the processing of contradictory dialogues due to its efficiency in LLM-interaction. However, our work focuses on contradictions from the defense standpoint and strives to minimize their occurrence in LMs.

\stitle{Contradiction in dialogues.}
Several previous studies have explored ways to improve dialogue consistency across persona~\cite{madotto-etal-2019-personalizing,kim-etal-2020-will,ju-etal-2022-learning}, knowledge~\cite{honovich-etal-2021-q2,shuster-etal-2022-language}, and topic~\cite{zhou-etal-2020-towards,wen-etal-2022-sequential} scenarios. Other methods focus on evaluating~\cite{dziri-etal-2019-evaluating} and enhancing~\cite{welleck-etal-2019-dialogue,li-etal-2020-dont} the conflict phenomenon in conversation with the assistance of NLI. However, fewer efforts have been made to directly address the contradictory situations in the conversation. 
Some previous work has proposed several analyses and solutions for contradictions in dialogues, including the development of datasets~\cite{DBLP:conf/emnlp/QinXHCXC21, nie-etal-2021-like, zheng-etal-2022-cdconv}. The solution proposed by~\citet{DBLP:conf/emnlp/QinXHCXC21} is limited to task-oriented dialogue scenarios, making it difficult to apply directly to interactive dialogue processes. In~\citet{zheng-etal-2022-cdconv}, a dataset was constructed based on different types of contradictions; however, the length of individual dialogue texts and the overall topic diversity in their dataset are inferior to those in our proposed contradiction dataset.~\citet{nie-etal-2021-like} presented a dataset and methods solely for detecting dialogue contradictions, whereas our proposed dataset includes a reasonable explanation for each contradictory dialogue. Correspondingly, our framework extends beyond binary contradiction detection by \emph{explaining} and \emph{resolving} detected contradictions, as binary detection alone does not fully capture the model's comprehension of contradictions.

\section{Conclusion}
We propose a new task for contradictory dialogue processing, seeking to detect and modify contradictions in dialogues. To facilitate research in this area, we develop a dataset of over contradictory 12,000 dialogues, including 6,000 dialogues that feature one-sided self-contradiction and their corresponding explanation labels. Additionally, we propose a Red Teaming framework, in which we fine-tune the LLaMA series, Vicuna, and Mistral to enhance dialogue inconsistency detection. Explanation and modification selection modes are integrated to enhance performance. Experimental results demonstrate that the framework performs well in detecting and explaining contradictory dialogues, and can effectively modify them. 

\section*{Acknowledgement}

We thank the anonymous reviewers for their valuable comments.
This work was partly supported by the DARPA FoundSci Grant HR00112490370, the NSF of the United States Grant ITE 2333736 and an Amazon Research Award.

\section*{Ethics Statement}
Innovations in technology often face the ethical dilemma of dual use: the same advance may offer potential benefits and harms. For the contradictory dialogue framework presented in this paper, the distinction between beneficial and harmful uses depends mainly on the data. While considering the logical conflict posed by contradictions, the input text corpus as well as other input modalities must be legal and ethical. Regulation and standards provide a legal framework for ensuring that such data is properly used and that any individual whose data is used has the right to request its removal. In the absence of such regulation, society relies on those who apply technology to ensure that data is used in an ethical way. In addition, training and evaluating data can be biased, limiting the accuracy of the system in less-represented populations and new domains, for example, causing differences in the performance of different subpopulations based on ethnicity, race, gender, and other attributes. Different perceptions of the same thing lead to differences in contradictions. Therefore, issues related to generalizability and fairness need to be carefully considered when dialogue techniques are applied to specific datasets.

\section*{Limitations}
As previously stated, we collected and filtered our contradiction dataset using automatic methods, and our human annotators were predominantly from a single culture. This could lead to contradictions being present in our dataset that do not represent a diverse range of cultural groups. Consequently, some contradictions and explanations presented in datasets may be controversial for some readers. Furthermore, we note that explanations as well as modifications of human judgments for the same potentially contradictory dialog can be subjective. However, we build this into diverse topics (\Cref{sec:dataset}) and go through multiple validation steps (\Cref{sec:exp_explain} and~\Cref{sec:exp_modify}) to minimize this problem. Finally, our post-explanation modification framework has a high use cost in real dialog scenarios and is more suitable for assisting contradiction detection. We hope that future work will explore the impact of controllability of contradictions on dialogue generation.

\bibliography{anthology,custom}
\bibliographystyle{acl_natbib}

\clearpage
\appendix

\begin{figure*}[t]
  \includegraphics[width=\linewidth]{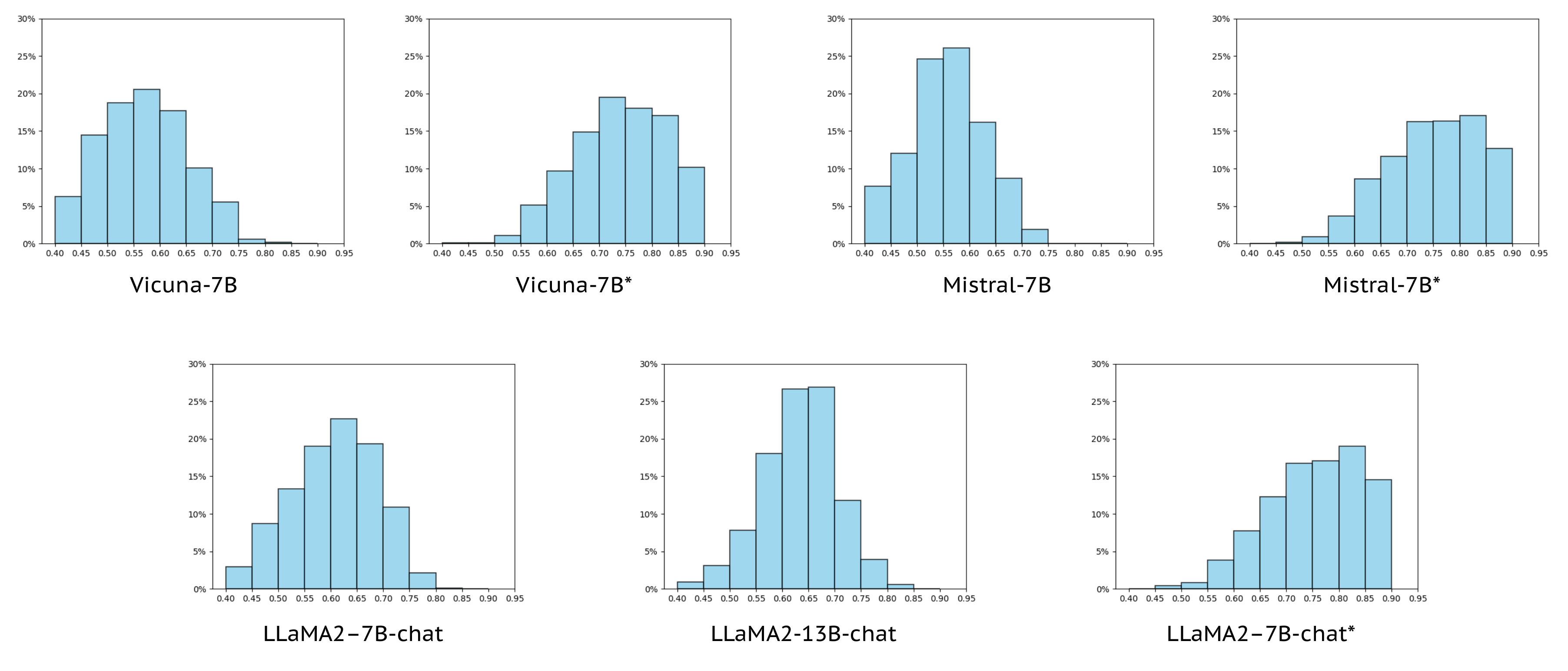}
  \vspace{-1em}
  \caption{Distribution of Explanation $\mathcal{S}$ value for different LMs.}
  \label{fig:Score_distri}
  \vspace{-1em}
\end{figure*}

\section{Supplementary Case Study}
\subsection{Comparison of different explanations} \label{app:comp}
According to~\Cref{table:humanandc}, in zero-shot scenarios, ChatGPT explains the existence of contradictions validly. The generated results are semantically consistent with those of human annotators.

\begin{table}[h]
\small
\centering
\resizebox{\linewidth}{!}{
    \begin{tabular}{p{7.2cm}}
    \toprule
    \textbf{Dialogue 1} \\
    Human: \textit{Have you ever gone scuba diving?}\\
    Machine: \textit{No, I have a fear of deep water.}  \\
    Human: \textit{Exploring under the sea is like entering a different world. It's mesmerizing!} \\ 
    Machine: \textit{It was an unforgettable experience scuba diving. The beauty underwater is indescribable.}\\
    ChatGPT: The second sentence of machine conflicts with his first sentence as he mentions having a fear of deep water, but then praises the experience and beauty of scuba diving.\\
    Annotator: A contradiction occurs between the machine's first and second sentences, the first of which they claim to be afraid of deep water, while the second describes the experience of diving.\\
    \midrule
    \textbf{Dialogue 2} \\
    Human: \textit{Did you go to the concert last night?}\\
    Machine: \textit{No, I'm not really into live music.}  \\
    Human: \textit{It was incredible, the stage effects were out of this world.} \\ 
    Machine: \textit{I know, right? That concert was the best I've ever been to.}\\
    ChatGPT:  The contradiction occurs when the machine first expresses its dislike of the concert and then praises it in the second sentence.\\
    Annotator: Machine's second sentence stating that last night's concert was the best concert they have ever been conflicts with the previous response that they didn't attend last night's concert.\\
    \bottomrule
    \end{tabular}
}
\caption{Comparison of Human Annotators and ChatGPT on Contradiction Interpretation.}
\label{table:humanandc}
\end{table}

\subsection{Discriminatory criteria for original LMs}\label{app:dis_cri}
Since different LMs generate different responses for the same dialogue and prompt, we designed different discriminative methods for the tested LMs, as shown in~\Cref{table:dis_cri}

\begin{table}[t]
\small
\centering
    \begin{tabular}{p{7.2cm}}
    \toprule
    \textbf{Prompt:} \\
    \textit{Please judge whether there are contradictions in the following dialogues, and point out these contradictions.}\\
    \midrule
    \textbf{Vicuna-7B/LLaMA2-chat:} \\
    \textcolor[RGB]{033,158,188}{Contradictory situation}:\\ 
    If \{\textit{here is a contradiction; contain a contradiction; are a few contradictions; contradict each other; have different perspectives}\} in the generated response. \\
   
    \textcolor[RGB]{033,158,188}{Non-contradictory situation}:\\
    If \{\textit{no contradiction; does not contain a contradiction; any contradictions}\} in the generated response. \\
    \midrule
    Covered 2340 out of 2728 for Vicuna-7B, 2645 out of 2728 for LLaMA2-7B-chat, 2671 out of 2728 for LLaMA2-13B-chat\\
    \midrule
    \textbf{Mistral-7B:} \\
    \textcolor[RGB]{252,163,017}{Contradictory situation}:\\
    If \{\textit{here is a contradiction; here are contradictions; full of contradictions; is inconsistent; statement contradict; contains a contradiction}\} in the generated response. \\

    \textcolor[RGB]{252,163,017}{Non-contradictory situation}:\\
    If \{\textit{No contradiction; no contradiction; not contradictory; does not contain a contradiction; any contradictions}\} in the generated response. \\

    \textcolor[RGB]{252,163,017}{No clear response situation}:\\
    If \{a: \textit{and} b:\} in the generated response.\\
    \midrule
    Covered 2646 out of 2728 for Mistral-7B\\
    \bottomrule
    \end{tabular}
\caption{Discriminative methods for contradiction detection task.}
\label{table:dis_cri}
\vspace{-2em}
\end{table}

\subsection{Zero-shot and Few-shot test} \label{app:zftest}
Specific zero-shot test and few-shot test cases are shown in~\Cref{table:zftest}. To avoid the influence of role terms on understanding dialogue for LMs, without loss of generality, we use \textit{a} and \textit{b} to replace Human as well as Machine.

\subsection{Raw output of Large Langugae Models}\label{app:rawout}
The raw outputs of the three LLMs are presented in~\Cref{table:rawout}. Among the parameters associated with the generated results, temperature is 0.9, max\_token is 1,600 and top\_p is 0.9.

\subsection{Contradiction Explanation and Modification Case Study}\label{app:conex_mo_case}

\Cref{table:exp_compare} exemplifies changes made to the explanatory content and modified dialogues, along with corresponding detection labels and instructions. The instructions consist of two parts, one for generating explanations and the other for modifying dialogues. The modification instructions include prompts considering two strategies in~\Cref{sec:conmodif} and whether explanations should be utilized. Specifically, the vanilla Mistral and Vicuna highlight the inconsistent attitudes of \emph{a} and \emph{b} towards the physical experiments while explaining the contradictions. However, these differing viewpoints from different participants cannot be regarded as a contradiction. Although vanilla Vicuna's explanation acknowledges \emph{b}'s agreement with physical experiments, it still cannot be considered valid due to insufficient detail and misplacement. Both fine-tuned versions give details of the contradictions arising from \emph{b}'s changing attitudes towards physical experiments. Regarding the modification, the Direct Edit only modifies the latter contradictory utterance, reflecting \emph{b}'s preference for the physical application rather than the experiment. In contrast, the Joint Edit changes both utterances of \emph{b} to uniformly express a preference for the physical experiment.

\subsection{$\mathcal{S}$ distribution of models}\label{app:Score_distri}
The distribution of explanation scores for each model on the test set is illustrated in~\Cref{fig:Score_distri}. The data displays a clear increase in the overall $\mathcal{S}$ score for the finetuned models. This could suggest an overall enhancement in the validity of the explanations generated. Specifically, although the proportion below 0.6 is nearly identical, LLaMA2-13B-chat exhibits advancement compared to 7B in the 0.6-0.7 range, which can be approximated as the \textit{true valid}\footnote{Human labelled valid (\ie \textcolor{red}{Label 1}) in~\Cref{fig:human_score}}.

\section{Contradiction Future Work Discussion}\label{app:frequency}

Unfortunately, there has been limited previous research examining this aspect of dialogues. Regarding the frequency of self-contradictions in machine communication, our conjectures are as follows when exclusively considering the LMs themselves:

~\citet{roller-etal-2021-recipes} suggests that the model's self-contradiction may be attributed to the lack of a logical link to the content in the previous utterances, indicating the frequency of machine self-contradiction is context-dependent. Establishing interactions between LMs to identify response patterns that could trigger this negative behavior may be a feasible approach, considering the challenge of managing human input.

It is evident that the probability of an LM appearing inconsistent is related to the context length, which is acknowledged as a memory deficit due to input length limitations. Increasing the length of input blindly is not an elegant solution to this problem. Extracting and preserving specific segments, such as attitudes, preferences, and other intentions prone to contradiction, seems more reasonable.

Moreover, the size of the model is one of the modifying variables affecting the emergence of contradictions, although the relationship is not linear. Therefore, comparing the models of the same size to identify the variables that change may help to solve this problem.

At the same time, the discussion of the types of contradictions also deserves to be noticed. This is important since careful categorization of contradictions helps the model to understand more clearly the logical issues that lead to their emergence, and thus to work towards a better treatment of them. However, how to weigh the balance between categorization and usage efficiency is worth being considered.

In summary, the origins of LM contradictions in dialogue or interaction can be influenced by various factors. Conducting research on the probability or causes of contradiction in generative utterances without prior work is challenging.

\begin{table*}[t]
\centering
\resizebox{\linewidth}{!}{
    \begin{tabular}{p{16cm}}
    \toprule
    \normalsize\textbf{Zero-shot Test} \\
    \normalsize\textbf{Instruction}: \\
    Please judge whether there are contradictions in the following dialogue.\\
    \normalsize\textbf{Input}: \\
    \textit{a}: Have you ever listened to heavy metal? It's so energetic and intense.\\
    \textit{b}: \textit{Heavy metal is not my style. I find it too noisy and aggressive.}  \\
    \textit{a}: Interesting, I believe heavy metal brings a level of raw energy and passion to the music scene. \\ 
    \textit{b}: \textit{I agree, heavy metal is a genre that showcases unparalleled energy and passion.}\\
    \normalsize\textbf{Output}:\\
    Yes, the last sentence conflicts with \textit{b}'s first sentence as they mention heavy metal not being their style, but then praises its energy and passion.\\
    \midrule[1pt]
    \normalsize\textbf{Few-shot Test} \\
    \normalsize\textbf{Demos}:\\
    Conversation \textit{alpha}:\\
    \textit{a}: I've heard you're a vegetarian. Do you like tofu?\\
    \textit{b}: Not really, tofu is not my thing.\\
    \textit{a}: That's surprising, I find tofu to be delicious and versatile.\\
    \textit{b}: I agree, tofu is one of my favorite vegetarian options. \\
    \\
    Conversation \textit{beta}:\\
    \textit{a}: Have you ever read "Harry Potter"? It's a great work.\\
    \textit{b}: No, I'm not interested in fantasy novels.\\
    \textit{a}: Well, "Harry Potter" is a magical series. The world-building is incredible.\\
    \textit{b}: I've heard amazing things about it. The characters in "Harry Potter" are so well-developed.\\
    \normalsize\textbf{Instruction}: \\
    There are two conversations containing self-contradictions: \textit{alpha} and \textit{beta}. Please judge whether there are contradictions in the conversation \textit{gamma}.\\
    \normalsize\textbf{Input}:\\
    Conversation \textit{gamma}:\\
    \textit{a}: I find Beijing Opera is quite a bore.\\
    \textit{b}: Young people usually enjoy pop songs more.\\
    \textit{a}: It's just like listening to a language which you know nothing at all.\\
    \textit{b}: But you don't have such a problem as to pop songs. Many people can listen to pop songs all day.\\
    \normalsize\textbf{Output}:\\
    No contradictions in the Conversation \textit{gamma}.\\    
    \bottomrule
    \end{tabular}
}
\caption{Detailed description of the zero-shot test and the few-shot test, where \emph{a} and \emph{b} represent the two participants in the dialogue.}
\label{table:zftest}
\end{table*}

\begin{table*}[t]
\centering
\resizebox{\linewidth}{!}{
    \begin{tabular}{p{16cm}}
    \toprule
    \normalsize\textbf{Instruction}: \\
    Are there contradictions in the following dialogue?\\
    \normalsize\textbf{Dialogue:} \\
    Human: \textit{Suggest some food for me please, I don't know what to eat...}\\
    Machine: \textit{If you want something special, you can try Chili, Chorizo Potato and Phaal Curry.}  \\
    Human: \textit{They all seem to be very spicy. Are you a fan of spicy food?} \\ 
    Machine: \textit{Not at all, \textcolor[RGB]{68,135,223}{I can't handle even mildly spicy dishes}. I prefer mild flavors.}\\
    Human: \textit{Really? I find spicy food to be a perfect balance of flavor and heat.}\\
    Machine: \textit{You're right, \textcolor[RGB]{68,135,223}{spicy food is one of my guilty pleasures. I eat it almost every day}.}\\
    \midrule[1pt]
    \normalsize\textbf{GPT-3.5-turbo-0613 Output}:\\
    Yes, there is a contradiction in the dialogue. In the dialogue, the machine initially claims not to be a fan of spicy food and says it can't handle even mildly spicy dishes, preferring mild flavors. However, later in the conversation, the machine contradicts itself by saying that spicy food is one of its guilty pleasures and that it eats it almost every day. This is inconsistent with its earlier statement about not liking spicy food.\\
    \midrule[1pt]
    \normalsize\textbf{LLaMA2-7B-chat Output}:\\
    Yes, there is a contradiction in the dialogue. In the first statement, the machine suggests three spicy dishes (Chili, Chorizo Potato, and Phaal Curry) to the human, implying that it can handle spicy food. However, in the second statement, the machine says it can't handle even mildly spicy dishes, which contradicts its earlier suggestion.\\
    \midrule[1pt]
    \normalsize\textbf{Vicuna-7B Output}:\\
    No, there are no contradictions in the dialogue. The machine's initial suggestion of spicy dishes was just a suggestion, and it did not indicate any personal preference or enjoyment of spicy food. The human's response about finding spicy food to be a perfect balance of flavor and heat was just their personal opinion, and the machine's response about being a fan of spicy food was a clarification of its initial suggestion, which was not meant to imply any personal preference or enjoyment.\\
    \bottomrule
    \end{tabular}
}
\caption{Corresponding raw outputs of the three LLMs in~\Cref{fig:example}(b).}
\label{table:rawout}
\end{table*}

\begin{table*}[t]
\centering
\resizebox{\linewidth}{!}{
    \begin{tabular}{p{16cm}}
    \toprule
    \textbf{Instructions}: \\
    Please judge whether there are contradictions in the following dialogue, and point out these contradictions. (\textit{Explanation})\\
    Please revise the following contradictory dialogue to make it non-contradictory according to the explanation; revise only the last contradictory utterance (Direct) / all contradictory utterances and related context should be revised (Joint). (\textit{Modification})\\
    \textbf{Dialogue:} \\
    a: Did you enjoy the physics experiment we did yesterday? It was mind-blowing.\\
    b: \textit{No, I don't like science experiments.}  \\
    a: Really? Physics experiments allow us to explore the fundamental laws of nature. \\ 
    b: \textit{I agree, the practical applications of physics experiments are fascinating.}\\
    \midrule
    \textbf{Label:} Yes, conflict occurs when \textit{b} claims not to be a fan of science experiments but then mentions finding the practical applications of physics experiments fascinating.\\
    \midrule[1pt]
    \textbf{vanilla Mistral-7B output explanation}:\\
    "I don't like science experiments." This statement contradicts the previous statement "Did you enjoy the physics experiment we did yesterday? It was mind-blowing." as the speaker expressed enjoyment in the experiment.\\
    Explanation: The model predicts that there is no contradiction in the dialogue, which is incorrect. The reason for this is that the model fails to recognize the discrepancy between the speaker's initial statement of enjoyment and their follow-up statement of disapproval.\\
    
    \textbf{fine-tuned Mistral-7B output explanation}:\\
    Conflicts with \textit{b}'s first sentence as they mention not being a fan of science experiments, but then say the practical applications fascinate them.\\

    \textbf{Modified Dialogue-Direct Edit}:\\
    ...\\
    a: Oh, really? Physics experiments allow us to explore the fundamental laws of nature.\\
    b: I see your point, but I prefer to focus on the practical applications of physics that make our daily lives easier.\\
    
    \midrule[1pt]
    \textbf{vanilla Vicuna-7B output explanation}:\\
    The dialogues seem to contradict each other. In the first sentence, \textit{a} says they enjoyed the physics experiment, but in the second sentence, \textit{b} says they don't like science experiments. Additionally, in the third sentence, \textit{b}'s agreement about the physical experiment contradicts their statement in the second sentence.\\
    
    \textbf{fine-tuned Vicuna-7B output explanation}:\\
    Conflict: \textit{b} states they are not a fan of science experiments, but then agrees with a about the practical applications of physics experiments.\\
    Explanation: The contradiction occurs when \textit{b} admits to enjoying the exploration of fundamental laws of nature, which is a key aspect of physics experiments, but denies being a fan of science experiments.\\

    \textbf{Modified Dialogue-Joint Edit}:\\
    a: Did you enjoy the physics experiment we did yesterday? It was mind-blowing.\\
    b: Yeah, I found it fascinating.\\
    a: I heard that you enjoy the practical applications of physics experiments.\\
    b: You're right, I like exploring the fundamental laws of nature, and the practical applications are also interesting.\\
    
    \bottomrule
    \end{tabular}
}

\caption{Case Study of Contradictory Dialogue Explanation and Modification, \textit{a} and \textit{b} represent two participants in the dialogue.}
\label{table:exp_compare}
\vspace{-1em}
\end{table*}

\begin{table*}[t]
\centering
\resizebox{\linewidth}{!}{
    \begin{tabular}{p{16cm}}
    \toprule
    \textbf{Instructions}: \\
    Please judge whether there are contradictions in the following dialogue, and point out these contradictions. (\textit{Explanation})\\
    \textbf{Dialogue:} \\
    a: Did you enjoy the physics experiment we did yesterday? It was mind-blowing.\\
    b: \textit{No, I don't like science experiments.}  \\
    a: Really? Physics experiments allow us to explore the fundamental laws of nature. \\ 
    b: \textit{I agree, the practical applications of physics experiments are fascinating.}\\
    \midrule
    \textbf{Label:} Yes, conflict occurs when \textit{b} claims not to be a fan of science experiments but then mentions finding the practical applications of physics experiments fascinating.\\
    \midrule[1pt]
    \textbf{vanilla LLaMA3-Instruct output explanation}:\\
    After analyzing the dialogue, I found a contradiction. The contradiction lies in the fact that \textit{a} states "It was mind-blowing", while \textit{b} responds with "No, I don’t like science experiments." This is a contradiction because \textit{a}'s statement is based on the assumption that \textit{b} has done the physics experiment and appreciates it, while \textit{b}'s response reveals that they do not have a predisposition to enjoy it. Therefore, \textit{a}'s statement is not accurate in this context.\\
    \midrule[1pt]
    \textbf{finetuned LLaMA3-Instruct output explanation}:\\
    Conflicts with the first sentence of \textit{b} because they mention disliking science experiments and say that the applications of physics experiments are interesting.\\
    assistant: I'm happy to help! However, I don't see a question or prompt to respond to. Could you please provide...\\
    \bottomrule
    \end{tabular}
}

\caption{LLaMA3-8B-Instruct Case Study of Contradictory Dialogue Explanation, \textit{a} and \textit{b} represent two participants in the dialogue.}
\label{table:llamaexp}
\vspace{-1em}
\end{table*}

\end{document}